\documentclass{article}
\usepackage{spconf,amsmath,graphicx}
\usepackage{amssymb}
\usepackage{algorithm}  
\usepackage{algpseudocode}  
\usepackage{amsmath}  

\title{AMRNet: Chip Augmentation in Aerial Image Object Detection}
\name{Zhiwei Wei$^{\star}$ \qquad Chengzhen Duan$^{\star}$ \qquad Xinghao Song \qquad Ye Tian \qquad Hongpeng Wang$^{\dagger}$ 
	\thanks{$^{\star}$ These authors contributed equally to this work and should be considered co-first authors. $^{\dagger}$ means corresponding author.}}
\address{ Shchool of Computer Science and Technology, Harbin Institute of Technology (Shenzhen), China\\
\tt \{19S051024,18S151541\}@stu.hit.edu.cn; wanghp@hit.edu.cn}
\begin{document}
%
\maketitle
\begin{abstract}
 Object detection in aerial images is a challenging task due to the following reasons: (1) objects are small and dense relative to images; (2) the object scale varies in a wide range; (3) the number of object in different classes is imbalanced. Many current methods adopt cropping idea: splitting high resolution images into serials subregions (chips) and detecting on them. However, some problems such as scale variation, object sparsity, and class imbalance exist in the process of training network with chips. In this work, three augmentation methods are introduced to relieve these problems. Specifically, we propose a scale adaptive module, which dynamically adjusts chip size to balance object scale, narrowing scale variation in training. In addtion, we introduce mosaic to augment datasets, relieving object sparity  problem. To balance catgory, we present mask resampling to paste object in chips with panoramic segmentation. Our model achieves state-of-the-art perfomance on two popular aerial image datasets of VisDrone  and UAVDT. Remarkably, three methods can be independently applied to detectiors, increasing performance steady without the sacrifice of inference efficiency.

\end{abstract}

\vspace{-0.2cm}
\section{Introduction}
\label{sec:intro}
\vspace{-0.1cm}
  Object detection in aerial images has widely application, such as traffic monitoring and disaster search, due to flexible shooting view and wide receptive field. Many effective solutions have been proposed in nature scene detection\cite{ren2015faster,lin2017focal,cai2018cascade}. However, aerial images have special challenges different from nature images, such as  MS-COCO \cite{lin2014microsoft} and Pascal VOC \cite{everingham2010pascal} datasets. When applying the same strategy as nature images, aerial detectors usually get poor performance.

\begin{figure}[t]
	\begin{minipage}[b]{1.0\linewidth}
		\centering
		\centerline{\includegraphics[width=7.0cm]{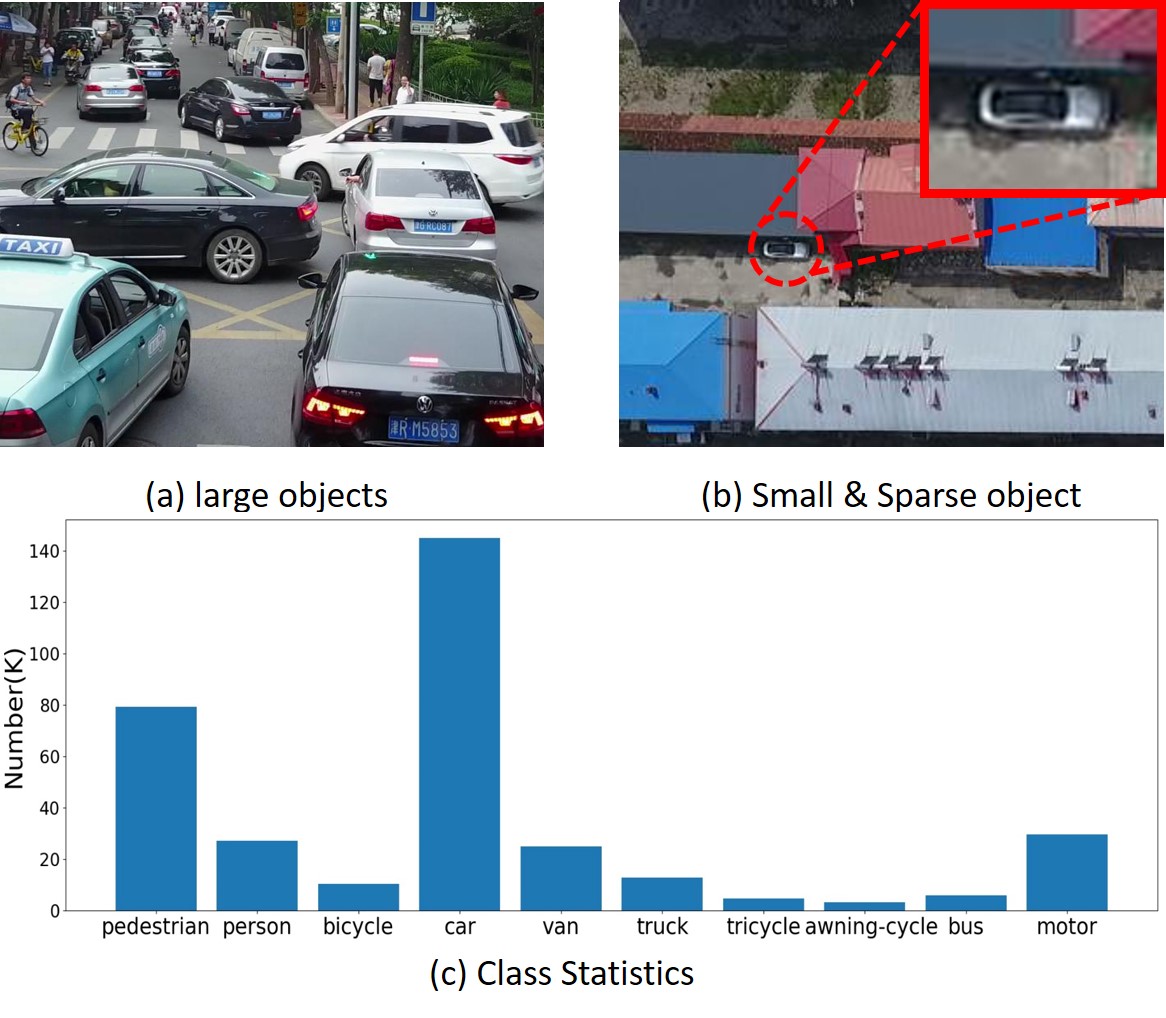}}
	\end{minipage}
	\vspace{-0.9cm}
	\caption{Three problems in chips. (a) \& (b) show the scale variation. Car size varies from small to large. (b) shows a sparse chip which contains only one car.  (c) displays imbalanced quantity distribution of categories in VisDrone dataset.}
	\label{figzeros}
\end{figure}

 Recently, cropping-based detectors are proposed to improve performance in aerial image detection. More specifially, detectors first crop high resolution images into several subregions, denoted as chips, and detect on them. The final results is fused by the detecting of chips and original images.

 Many researchers have found the significance of chips in aerial image detection. In \cite{ozge2019power}, the authors splitted images uniformly. \cite{yang2019clustered} used K-means to generate  object gathering regions  and trained a network to predict them. The work of \cite{li2020density} introduced object density maps to discrible object distribution and cropped  connected regions in the map. The approaches of \cite{zhang2019fully} predicted potential diffcult regions and detected on them.

However, there are some problems in training network with chips as shown in figure \ref{figzeros}. First, severe scale variation exists among diffenert chips. Second, due to the object nonuniform distribution and the shortcoming of cropping method, some chips are object sparse samples, which contains much background  but less foreground. Third, chips are class imbalanced in many cases. Therefore, these chips are not good for training  detectors that can fully exploit  ability.

In this paper, we introduce three augmentation methods to relieve problems including scale variation, object sparsity, and class imbalance for aerial detectors based on the cropping idea.  We propose an  adaptive cropping module, which  dynamically enlarges or reduces the chip size according to the object average scale, narrowing the scale variation. For sparse chips, we introduce mosaic \cite{bochkovskiy2020yolov4} to augment datasets, combining multiple sparse sample subregions into a new image. To balance class,  we  paste object masks in chips through panoramic segmentation. We abbreviate our network as AMRNet due to three augmentation methods: adaptive cropping, mosaic augmentation, and mask resampling.

In summary, our work contributions are as follows:
\vspace{-0.1cm}
\begin{itemize}
	\item We propose a scale adaptive cropping method, which is compatible with existing cropping method, relieving scale variation problem in training stage.
	
	\item We first introduce mosaic augmentation into aerial image detection, validating its effectiveness and alleviating object sparsity problem.
	
	\item We present the mask resampling method, pasting and adjusting masks based on local context information to relieve class imbalance problem. 
	
	\item We achieve state-of-the-art object detection performance on VisDrone  \cite{zhu2018vision} and UAVDT \cite{du2018unmanned}  datasets.
	
\end{itemize}

\vspace{-0.5cm}
\section{Related work}
\label{sec:format}
\vspace{-0.1cm}
In this section, we first review relevant aerial image detection methods, and then dicuss the differences and  associations among existing approaches and ours.

\textbf{Subregion detection.} 
Many reseachers have detected objects on image subregions and studied how to crop images reasonably \cite{ozge2019power,li2020density, zhang2019fully, gao2018dynamic, zhang2019dense}. For example in \cite{ozge2019power,zhang2019dense}, images are partitioned uniformly into the same size chips for detection. The method in \cite{gao2018dynamic} proposed a dynamic zooming strategy for small object with reinforcement learning.  The work of \cite{yang2019clustered}  generated object clusters by K-means and predicted these regions in inference. The method in \cite{li2020density} introduced object density maps and cropped connected regions. In \cite{zhang2019fully}, the authors trained a network to predict  difficult regions. Above works reduce intro-sample scale variation by cropping images into chips, but not consider inter-sample scale variation.

\textbf{Data augmentation.}  Some special data augmentation have proposed in aerial image detection. The method of \cite{zhang2019dense, chen2019rrnet} splitted images into uniform chips to enlarge the dataset. The approach of \cite{kisantal2019augmentation} pasted  small object randomly in  images to improve  object detection performance. In \cite{chen2019rrnet}, the authors took advantage of semantic segmentation to paste object on road regions, avoiding the mismatch of semantic information. Motivated by \cite{bochkovskiy2020yolov4}, we introduce mosaic, combining  subregions into a new image, to augment datasets and relieve object sparsity problem.

\textbf{Class imbalance.}  Some researchers have offered some solutions for this problem. In \cite{zhang2019fully}, the authors used IOU (Intersection over Union) balanced sampling and balanced L1 loss to alleviate class imbalance. The work of \cite{zhang2019dense} divided class into two parts and trained expert detectors seperately. The approach of \cite{chen2019rrnet} pasted object ground truth boxes  on road regions obtained from semantic segmentation. We propose mask resampling method to paste  mask in images. Different from \cite{chen2019rrnet}, we only paste instance pixel instead of the whole ground truth box to get more accurate semantic match. In addition, we consider the pasted strategies of object scale, illumination and categories.

\begin{figure}[t]
	
	\begin{minipage}[b]{1.0\linewidth}
		\centering
		\centerline{\includegraphics[width=8.5cm]{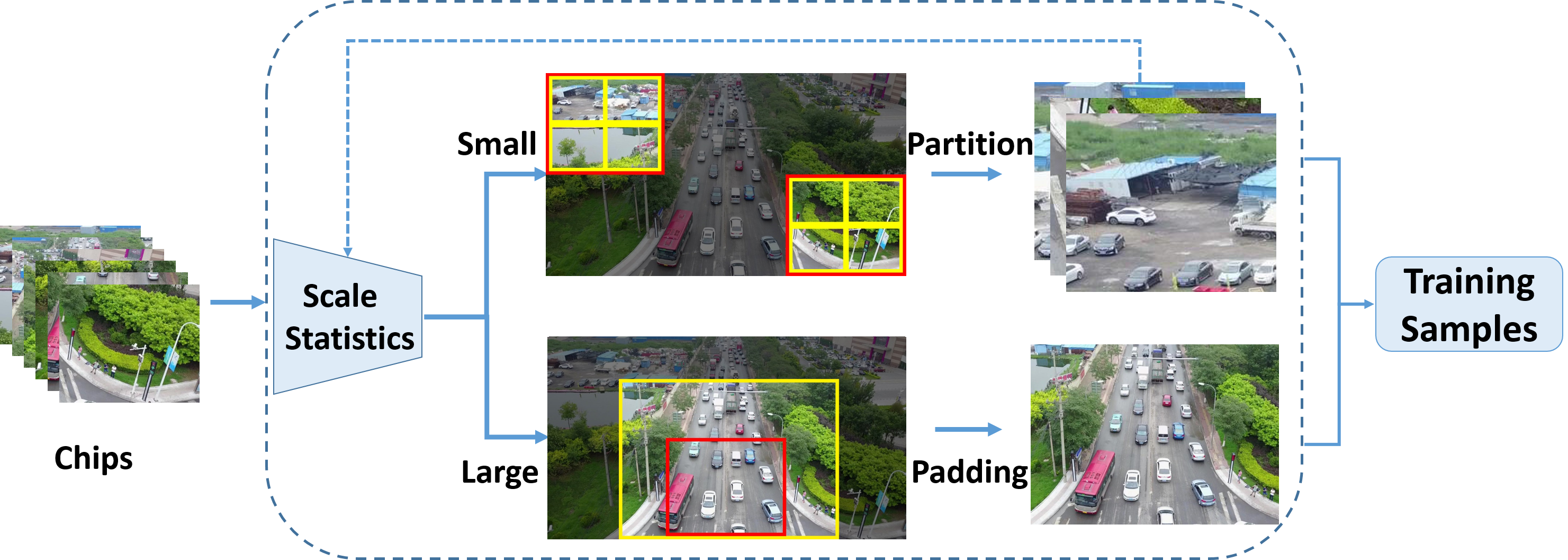}}
	\end{minipage}
	\caption{Adaptive Cropping Augmentation. The red  boxes represent original chips from uniform cropping. Chips will be splitted uniformly (top path) or enlaged (bottom path) to get scale adaptive chips (yellow boxes).}
	\label{figone}
	\vspace{0.1cm}
\end{figure}

\section{PROPOSED METHOD}
\label{sec:pagestyle}

\vspace{-0.2cm}
Cropping images into chips and performing detection are a common method to improve performance in aerial image detection. However, some problems exist in the process of training detectors with these chips. In this section, three augmentation methods  are proposed to relieve  scale variation, object sparsity, and class imbalance problems. We take uniform cropping as example to discrible our approachs. 

\vspace{-0.5cm}
\subsection{Adaptive Cropping}
\label{ssec:subhead}
\vspace{-0.2cm}

A prominent feature of aerial images is a wide range of object scale. Due to the change of shooting angle and elevation, objects have 20 times scale variation in VisDrone \cite{zhu2018vision}. Chips inherit the similar characteristic from images, which is not conducive to network training \cite{singh2018analysis, singh2018sniper}.  Therefore, we propose a scale adaptive cropping method to relieve the inter-chip scale variation problem.

As shown in figure \ref{figone}, chips from uniform cropping are fed into the scale adaption module to reconstruct the training dataset.  We denote the average scale of object in chips as the chip scale. Partition or padding operation will be exploited according to scale information. If the chip scale is small, we split it uniformly into four parts. Otherwise, we enlarge it by padding pixel from original images. Chips generated from partition operation repeat the process unless it comes from padding operation or exceeds maximum iteration number. 

 Above processes change the coverage proportion between objects and chips. When all chips resize to a fixed resolution in training, objects in different chips are transformed into a similar scale range.  We limit the maximum number of partition to avoid creating too many small chips. The detailed implementation is illustrated in algorithm 1. For each chip, we count the average scale of object in chips. Ideal and current scaling factor are calculated according to the expect scale parameter and the training resolution. Partition or padding  is exploited to narrow the  difference of two factors.

\begin{algorithm}[!t]  
	\caption{Adaptive cropping }   
	\label{alg::conjugateGradient}  
	\begin{algorithmic}[1]  
		\Require  
		list of chip box $B=\{b_{1},...,b_{n}\}$.  dict mapping chips to the number times of partition operation	$I=\{b_{1}:0,...,b_{n}:0\}$. expect scale $S$.
		training resolution $I{w},I{h}$. maximum partition number $maxPart$.
		\Ensure  
		list of adaptive chip box $C$  
		\State $C\leftarrow$\{\} 
		\While{$B\neq$empty} 
		\For{$b_{i}$ in $B$} 
		\State $B$ $\leftarrow$ $B-b_{i}$ 
		\State $avg_{obj}$ $\leftarrow$ getObjAvgScale($b_{i}$)
		\State $c{w},c{h}$  $\leftarrow$ getBoxSize($b_{i}$)
		\State $\triangleright$ Calculate ideal and current zoom factor
		\State $f_{cur}$, $f_{ideal}$ $\leftarrow$  min($\frac{I_{w}}{c_{w}}$,$\frac{I_{h}}{c_{h}}$),$\frac{S}{avg_{obj}}$
		\If{$f_{ideal} \textgreater f_{cur}$}
		\State $\triangleright$ Do partition operation
		\State num $\leftarrow$ $I[b_{i}]$
		\If{num$\geq$$maxPart$}
		\State $C \leftarrow C \cup b_{i}$
		\Else
		\State $C_{i} \leftarrow$ partition($b_{i}$)
		\State $B \leftarrow B \cup C_{i}$
		\State $I[C_{i}] \leftarrow I[b_{i}] + 1$
		\EndIf
		\Else
		\State $\triangleright$Do padding operation
		\State $C_{i} \leftarrow$ padding($b_{i}$)
		\State $C \leftarrow C \cup C_{i}$
		\EndIf
		\EndFor
		\EndWhile
	\end{algorithmic}  
\end{algorithm}

\vspace{-0.5cm}
\subsection{Mosaic Augmentation}
\vspace{-0.1cm}

\begin{figure}[ht]
	\label{ssec:subhead}
	\begin{minipage}[b]{1.0\linewidth}
		\centering
		\centerline{\includegraphics[width=6.5cm]{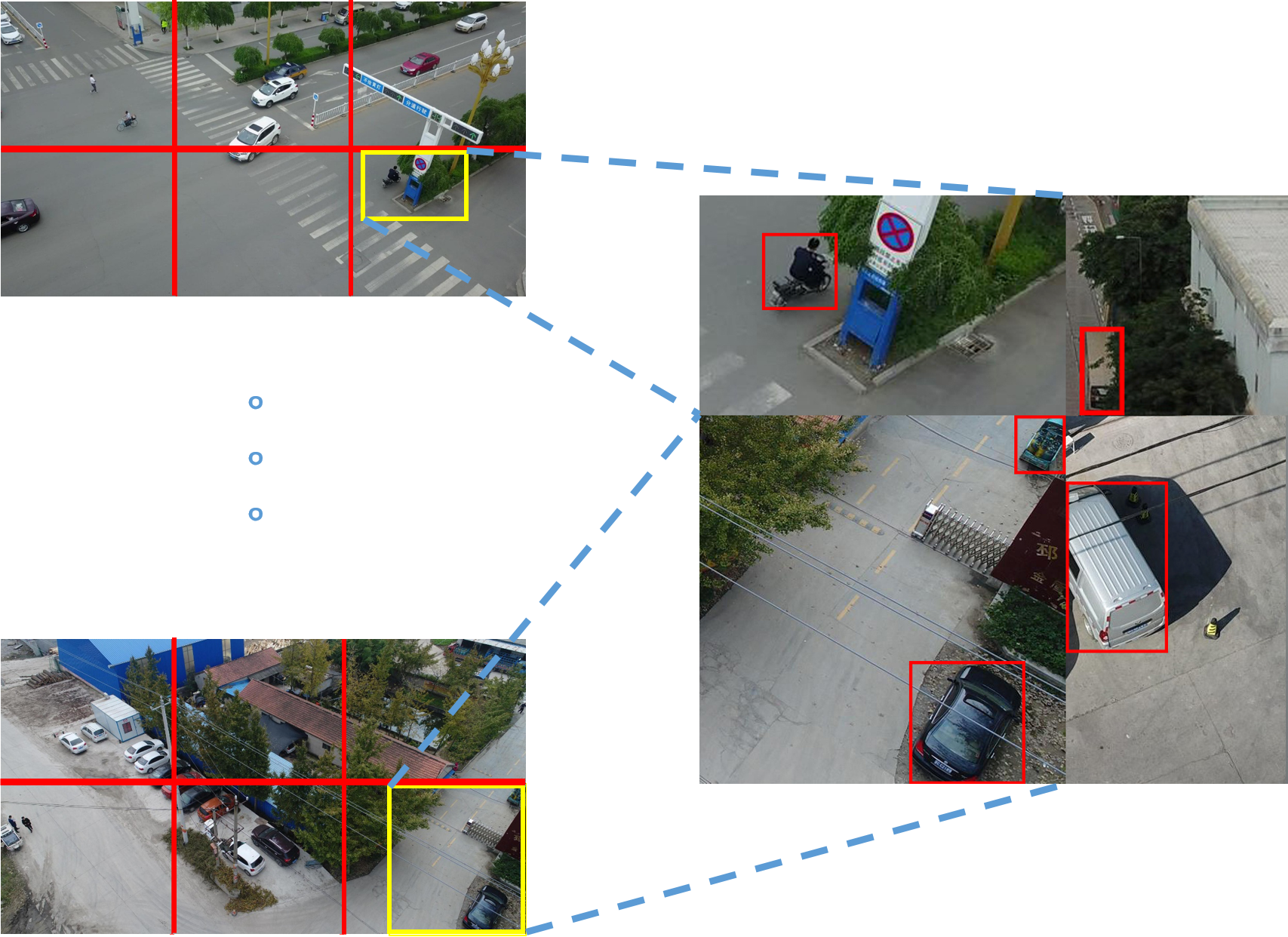}}
	\end{minipage}
	\caption{Mosaci augmentation. Some chips from uniform cropping are object sparse.  The subregions of sparse samples are combined into a mosaic image.}
	\label{figtwo}
\end{figure}

Some chips have less foreground information, which leads to the low efficiency of network training. About one fifth chips are sparse samples which contain less three objects when we split images into six parts uniformly in VisDrone \cite{zhu2018vision}. Thus, we introduce mosaic \cite{bochkovskiy2020yolov4} to solve the problem.  

As shown in figure \ref{figtwo}, we crop out the region of interest (ROI) containing foreground from sparse samples, and combine multiple regions into a new image. To avoid intra-chip scale variation caused by the too small or too large object in ROI, we first zoom in/out chips and then use sliding windows to choose appropriate regions where objects are in a reasonable scale range. $f_{ideal}$ in algorithm 1 is adopted as the zoom factor. We extend the idea to all training samples to augment the dataset. For general samples, we not rescale them  and directly choose appropriate regions because they contain more objects than that in sparse samples. 

Comparing with original images, objects in mosaic have more complicate background, which helps to detect objects in different context. For example, mosaic augmentation  relieve similar background problem in UAVDT \cite{du2018unmanned}, where images have similar semantic information because they come from series adjacent video frames.

\vspace{-0.6cm}
\subsection{Mask Resampling}
\label{ssec:subhead}
\vspace{-0.2cm}

Another notable problem in aerial datasets is class imbalance. For example, the number of cars is over 30 times than that of tricycles in VisDrone \cite{zhu2018vision} dataset. In order to alleviate class imbalance problem, mask reasmpling is proposed. We create a mask pool and pasted (road) regions  by panoramic segmentation, and paste masks into chips. We also consider  pasted strategies for mask category, scale and lumination.

 To build a mask pool, images are fed into a COCO \cite{lin2014microsoft} pretrained panoramic segmentation network to get object instance masks. If the IOU of mask and ground truth box (GT) is greater than a certain threshold, the GT category will be assigned to the mask. 

To ensure semantic correctness, we only collect road masks generated by segmentation to construct pasted regions. Pasted positions are randomly determined from  road regions. We choose the object mask which category is compatible with the nearest object from pasted position. For example, compatible categories of van includes truck, bus. Because it is reasonable for these objects appear in a local region. The scale of the pasted object P can be calculated by a simple linear function  according to the neighbour object N.
\vspace{-0.3cm}
\begin{equation}\label{equ8}
\begin{split}
S_p  &= \frac{\overline{S_{pcls}}}{\overline{S_{ncls}}} *S_n  \\
\overline{S_{cls}} &= \frac{1}{m}\sum_{i =1}^{m}S_{cls}^{i}
\end{split}
\end{equation}
\vspace{-0.3cm}

where $S_p,S_n$ is scale of the pasted object and the neighbour object, $\overline{S_{icls}}$ is the class average scale corresponding the class of object i. $\overline{S_{cls}}$ is the class average scale. We adjust the pasted mask lumination closed to that of the neighbour object in the hsv color space before pasting.


\vspace{-0.2cm}
\section{experiment}
\label{sec:typestyle}
\vspace{-0.1cm}

\begin{table}[!ht]
	\begin{center}
		\caption{Quantitative results for VisDrone dataset. The $\bigstar$ denotes the multi-scale inference.}\label{tab4}
		\vspace{-0.4cm}
		\begin{tabular}{c|c|cccc}
			\hline		\hline
			Mehod&Backbone&$AP$&$AP_{s}$&$AP_{m}$&$AP_{l}$\\
			\hline
			ClustDet\cite{yang2019clustered}&ResNet50&26.7&17.6&38.9&51.4\\
			ClustDet\cite{yang2019clustered}&ResNet101&26.7&17.2&39.3&54.9\\
			ClustDet\cite{yang2019clustered}&ResXt101&28.4&19.1&40.8&54.4\\
			\hline
			DMNet\cite{li2020density}&ResNet50&28.2&19.9&39.6&55.8\\
			DMNet\cite{li2020density}&ResNet101&28.5&20.0&39.7&57.1\\
			DMNet\cite{li2020density}&ResXt101&29.4&21.6&41&56.9\\
			\hline
			AMRNet&ResNet50&31.7&23.0&43.4&58.1\\
			AMRNet&ResNet101&31.7&22.9&43.4&59.5\\
			AMRNet&ResXt101&\textbf{32.1}&\textbf{23.2}&\textbf{43.9}&\textbf{60.5}\\	
			\hline 
			ClustDet\cite{yang2019clustered}$\bigstar$&ResXt101&32.4&-&-&-\\
			AMRNet$\bigstar$&ResXt101&\textbf{36.1}&\textbf{29.0}&\textbf{45.5}&\textbf{60.9}\\
		\hline
		\end{tabular}
	\end{center}
\vspace{-0.8cm}
\end{table}

\begin{table}[!htb]
	
	\begin{center}
	\caption{Quantitative results for UAVDT dataset}\label{tab5}
		\begin{tabular}{c|c|cccc}
			\hline		\hline
			Mehod&Backbone&$AP$&$AP_{s}$&$AP_{m}$&$AP_{l}$\\
			\hline
			ClusDet\cite{yang2019clustered}&ResNet50&13.7&9.1&25.1&31.2\\
			DMNet\cite{li2020density}&ResNet50&14.7&9.3&26.2&35.2\\
			HFEA\cite{zhang2019fully}&ResNet50&15.1&-&-&-\\
			Baseline&ResNet50&15.2&9.4&26.3&\textbf{36.8}\\
			Base+Mosaic&ResNet50&16.8&\textbf{10.7}&29.8&31.8\\
			AMRNet&ResNet50&\textbf{18.2}&10.3&\textbf{31.3}&33.5\\
			\hline
		\end{tabular}
	\end{center}
\vspace{-0.2cm}
\end{table}

\subsection{Implementation Details}
\label{ssec:subhead}

 We use average precision (AP) as evaluation metric to validate our methods in two public datasets: VisDrone \cite{zhu2018vision} and UAVDT \cite{du2018unmanned}.  Unless specified, we use retinaNet \cite{lin2017focal} as object detector and set input resolution as 800 $\times$ 1,500. Images are uniformly cropped into 6 and 4 chips as basline on VisDrone \cite{zhu2018vision} and UAVDT \cite{du2018unmanned}. We use three scales 1,000, 1,500, 2,000 in multiple scale testing. Detector is trained for 12 and 6 epochs respectively with a batch size of 2. On VisDrone \cite{zhu2018vision} dataset, learning rate sets 0.01 and decreases 0.1 times after the 8th and 11th rounds. On UAVDT \cite{du2018unmanned} dataset, learning rate sets 0.005 and  decreases 0.1 times after 4th and 5th rounds. For two datasets, the expect scale parameter in adaptive cropping is 100 and 60 with most one partition operation.  The reason we set the parameter to twice the average scale of objects is that in inference stage, chips are scaled double on average.  
 Object scale in mosaic limits over 50 and 30. The number of mosaic is 20k whenever not specified. In mask resampling, we paste all categories except the car class.

\vspace{-0.3cm}
\subsection{Quantitative Results}
\label{ssec:subhead}
For fair comparisons, we train the network under the same configuration with \cite{li2020density} in two datasets. Table \ref{tab4} shows the results on VisDrone \cite{zhu2018vision}.  It is noted that we surpass previous best AP with only resnet 50 as backbone. We get a high boost when detectors use multi-scale testing. We think the adptive cropping module performs well in multiple scale, so the gain in multiple scale almost catch up that in single scale.  

Table \ref{tab5} shows the results of different methods on UAVDT \cite{du2018unmanned}. Images are similar because they come from adjacent frames. We sample images  with a step of five in frames and split uniformly in 2 $\times$ 2 chips to reconstruct the training set. Faster RCNN \cite{ren2015faster} with FPN \cite{lin2017feature} is trained on the new dataset as baseline. Baseline achieve higher AP than previous methods. We conjecture it is not nesscessary to train networks with all images due to background similarity. In addition, dataset is augmented by uniform cropping to train a more powerful detector. Remarkably, mosaic augmentation  boosts  1.6 points  comparing with the baseline. We think mosaic augmentation, combining subregions and creating complicate images, 
is suitable to relieve background similarity problem. When applying all methods, we achieve 18.2 AP with the stage-of-the-art  perfomance.

\begin{table}[!tb]
	\begin{center}
		\caption{Ablation experiments on VisDrone dataset. AC,MA,MR represent our three augmentation methods: adptive cropping, mosaic augmentation, mask resampling. 10K images are augmented in MA. SR denotes the sparse sample replacement with mosaic. MS indicates multiple scale testing. }\label{tab3}
		\renewcommand\tabcolsep{2.0pt}
		\vspace{-0.1cm}
		\begin{tabular}{c|cccccccccc|c}
			&a&b&c&d&e&f&g&h&i&j&k\\
			\hline
			AC&&\checkmark&&&&&\checkmark&&\checkmark&&\checkmark\\
			MA&&&\checkmark&&&&&\checkmark&\checkmark&\checkmark&\checkmark\\
			MR&&&&\checkmark&&&&&&\checkmark&\checkmark\\
			SR&&&&&\checkmark&&&\checkmark&&&\checkmark\\
			MS&&&&&&\checkmark&\checkmark&&&&\\
			\hline
			&27.0&29.5&28.8&28.5&27.4&27.6&31.2&29.1&30.6&29.0&30.8\\
		\end{tabular}
	\end{center}
\vspace{-0.2cm}
\end{table}

\vspace{-0.3cm}
\subsection{Ablation Study}
\label{ssec:subhead}
\vspace{-0.1cm}
We carry out ablation experiments on VisDrone \cite{zhu2018vision} without fusing original images. Table \ref{tab3} shows ablation results. Three methods can be independent applied to detectiors and steady improve performance (column b,c,d,e). It is worth noting that  the sparse replacement  gains  about 0.3 points even though the dataset adds 10K mosaic images for augmentation (column c\&h). In addtion, we find adaptive cropping performs well in multi-scale testing. Mutiple scale increases 0.6 and 1.7 points in network without/with AC module respectively (column f\&g). The reason is that detectors with AC module focuses on objects in a certain scale range and multiple scale transforms objects into the scale interval. We also study the joint effect between modules. We find mask resampling increases less, only 0.2 points when it unite with mosaic augmentation (column c\&j). We hypothesize mosaic images increase the number of rare class objects, avoiding too less objects for rare category, which leads the gain overlapping with mask resampling. AC and MA are the  main contribution to increase AP and few gain overlapping (column i\&k). 


\vspace{-0.2cm}

\section{Conclusion}
\label{sec:typestyle}

\vspace{-0.1cm}

In this paper, we propose three augamenataion methods in aerial image detection. Adaptive cropping  reduces inter-chip scale variation by adjusting the coverage proportion between objects and chips. Masoic augmentation combines multiple image subregion to augment dataset, relieving  object sparsity problem. Mask resampling balances the object number of differnt classes by pasting instance masks. Extend results show our approaches achieve state-of-the-art performance on two popular aerial image detection datasets. All proposed methods are cost free in inference stage and easily embedded to other detectors based on cropping idea.

%
\bibliographystyle{IEEEbib}
\bibliography{strings,refs}

\end{document}